\documentclass[10pt,twocolumn,letterpaper]{article}

\usepackage{iccv}
\usepackage{times}
\usepackage{epsfig}
\usepackage{graphicx}
\usepackage{amsmath}
\usepackage{amssymb}
\usepackage[ruled,vlined]{algorithm2e}
\usepackage{color}

\usepackage[pagebackref=true,breaklinks=true,letterpaper=true,colorlinks,bookmarks=false]{hyperref}

\iccvfinalcopy 

\def\iccvPaperID{669} 

\ificcvfinal\fi
\begin{document}

\title{Attribute-Graph: A Graph based approach to Image Ranking}

\author{Nikita Prabhu and R. Venkatesh Babu\\
Video Analytics Lab, SERC\\
Indian Institute of Science, Bangalore, India.\\
{\tt\small nikita@ssl.serc.iisc.in, venky@serc.iisc.in}}

\maketitle
\thispagestyle{empty}

\begin{abstract}
   We propose a novel image representation, termed Attribute-Graph, to rank images by their semantic similarity to a given query image. An Attribute-Graph is an undirected fully connected graph, incorporating both local and global image characteristics. The graph nodes characterise objects as well as the overall scene context using mid-level semantic attributes, while the edges capture the object topology. We demonstrate the effectiveness of Attribute-Graphs by applying them to the problem of image ranking. We benchmark the performance of our algorithm on the `rPascal' and `rImageNet' datasets, which we have created in order to evaluate the ranking performance on complex queries containing multiple objects. Our experimental evaluation shows that modelling images as Attribute-Graphs results in improved ranking performance over existing techniques.
\end{abstract}

\section{Introduction}
In a digital world of Flickr, Picasa and Google Image Search, ranking retrieved images based on their semantic similarity to a query has become a vital problem. Most content based image retrieval algorithms treat images as a set of low level features or try to define them in terms of the associated text \cite{Krapac10,wang09}. Such a representation fails to capture the semantics of the image. This, more often than not, results in retrieved images which are semantically dissimilar to the query.

Image processing and computer vision researchers to date, have used several different representations for images. They vary from low level features such as SIFT \cite{Lowe04}, HOG \cite{Dalal05}, GIST \cite{Oliva01} etc. to high level concepts such as objects and people \cite{Li10}. Since we want our image retrieval system to rank images in a way which is compatible with visual similarity as perceived by humans, it is intuitive to work in a human understandable feature space. When asked to describe an object or a scene, people usually resort to mid-level features such as size, appearance, feel, use, behaviour etc. Such descriptions are commonly referred to as the attributes of the object or scene. These human understandable, machine detectable attributes have recently become a popular feature category for image representation for various vision tasks \cite{Saleh13,Zheng14seg,Zhou14}. In addition to image and object characteristics, object interactions and background/context information form an important part of an image description. It is therefore, essential, to develop an image representation which can effectively describe various image components and their interactions. 
\begin{figure}[!t]
 \begin{minipage}{0.28\linewidth}
 \includegraphics[scale=0.26]{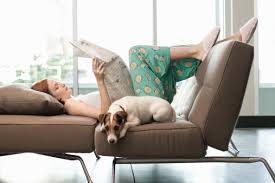}
 \centerline{(a)}
\end{minipage}
\hspace{6mm}
\begin{minipage}{0.28\linewidth}
 \includegraphics[scale=0.26]{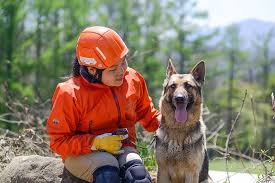}
 \centerline{(b)}
\end{minipage}
\hspace{6mm}
\begin{minipage}{0.24\linewidth}
 \includegraphics[scale=0.21]{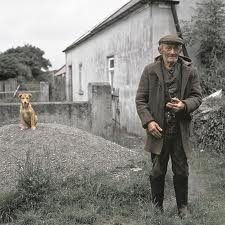}
 \centerline{(c)}
\end{minipage}
 \caption{Same objects, yet different semantics: (a) A dog next to a person lying on a sofa (b) A person and a dog near a forest (c) A dog watching a person from a distance}
 \label{fig:eg1}
\end{figure}
\begin{figure}
 \begin{minipage}{0.28\linewidth}
 \includegraphics[scale=0.26]{./Figures/query_eg1}
 \centerline{(a)}
\end{minipage}
\hspace{6mm}
\begin{minipage}{0.19\linewidth}
 \includegraphics[scale=0.23]{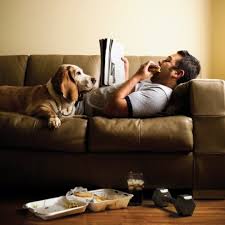}
 \centerline{(b)}
\end{minipage}
\hspace{6mm}
\begin{minipage}{0.26\linewidth}
 \includegraphics[scale=0.26]{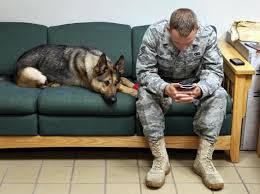}
 \centerline{(c)}
\end{minipage}
 \caption{(a) An example query image (b) \& (c) Expected Retrieved images: Ones with same semantics}
 \label{fig:eg2}
\end{figure}

\begin{figure*}[!t]
 \begin{minipage}{0.49\linewidth}
\includegraphics[scale=0.34]{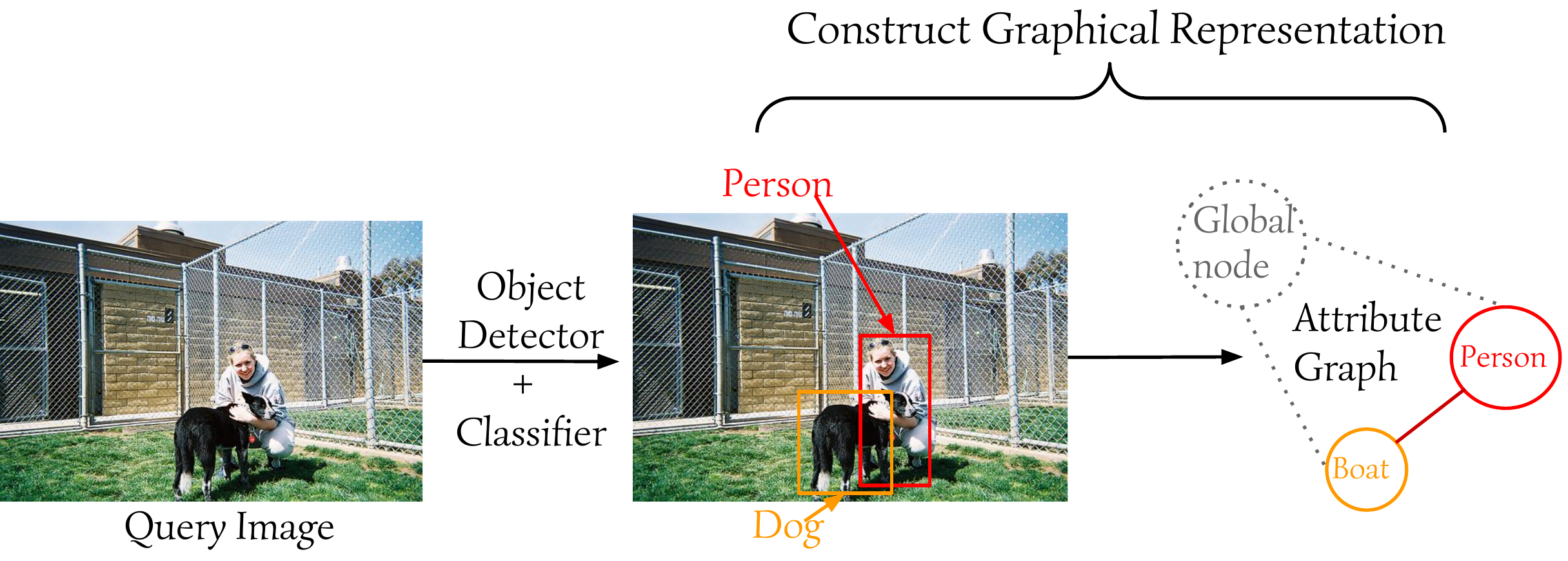}
\end{minipage}
\begin{minipage}{0.49\linewidth}
\includegraphics[scale=0.39]{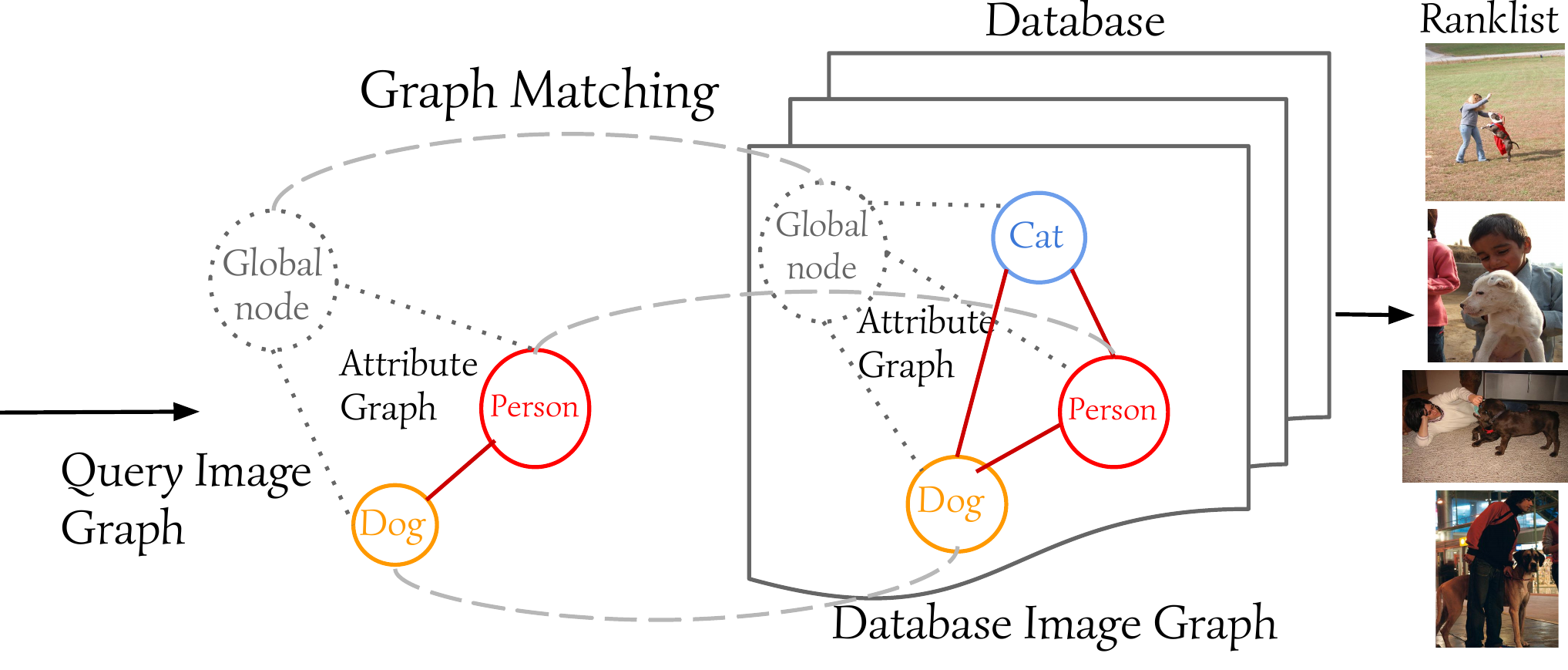}
\end{minipage}
 \caption{Overview of the proposed method: Object detection and classification is followed by the construction of the Attribute-Graph of the query. The query attribute-graph is then compared with the Attribute-Graphs of the database images, via graph matching to obtain a ranklist.}
 \label{fig:Overview}
\end{figure*}

Graphs, which have long been used by the vision community to represent structured groups of objects, are an ideal tool for this purpose~\cite{Felzenszwalb04,Jones14,Lu14,Shi97}. We represent images as Attribute-Graphs, using graph nodes to represent objects in the image and the global context of the scene. Each object is described using object-specific local attributes, and the overall scene with global attributes, thereby capturing both local and global descriptions of the image. Apart from the objects of importance in the scene, our model also incorporates the geometrical layout of the objects in the image. The edges of the graph capture the location, structure and orientation of the nodes with respect to other nodes. The proposed Attribute-Graphs thus characterise an image from three different perspectives, namely, global (scenes), local (objects) and sub-local (attributes) as well as the inter-relations between the same. This allows them to better conceptualise the essence of the image. 

We then use graph matching to ascertain the similarity of each of the dataset image models to the query model. The pipeline of the proposed method is shown in Fig.~\ref{fig:Overview}. The goal of our work is to rank images according to their semantic similarity to the query, by developing an image representation capable of unifying the varied aspects in an image.

The rest of the paper is organised as follows. In Section \ref{sec:rel_work} we detail the prior work in this area. In Section \ref{sec:prob_formulation}, we discuss crucial aspects of the concept of semantic similarity of images. Keeping the aforementioned aspects in mind,we describe the construction of Attribute-Graphs in Section \ref{sec:proposed_method}. In Section \ref{sec:dataset}, we describe our datasets `\textit{rPascal}' and `\textit{rImageNet}' created to evaluate image ranking.  The performance of the proposed algorithm is discussed in Section \ref{sec:results}. Section \ref{sec:conclusion} concludes the paper with a summary of our work.

\section{Related Work}
\label{sec:rel_work}

Mid level features have long been used for various tasks in the field of computer vision. The recent surge in the use of attributes have reiterated the efficacy of such features. Farhadi \textit{et al.}~\cite{Farhadi09} showed the usefulness of attributes for image description, object classification and abnormality detection. Extending the applicability of attributes beyond objects, Patterson \textit{et al.}~\cite{Patterson12} demonstrated their use for scene description and classification. Attributes have also been employed for image understanding \cite{patterson14,Hays13}, web search \cite{Kovashka13,Grauman13}, zero-shot-learning \cite{Jayaraman14,Lampert13}, action recognition \cite{sharma13} and human-computer interaction \cite{Parikh13}.

In addition to these applications, attributes have recently been increasingly used for the task of image retrieval and ranking. As attributes can effectively describe mid-level semantic concepts, they are a convenient means to express the users' search intentions. However most attribute based image retrieval works \cite{Siddiquie11, Yu12} focus on textual queries. For example, Siddiquie \textit{et al.}~\cite{Siddiquie11} exploit the interdependence among both query and non-query attributes, and model the correlation among them to improve retrieval performance. Yu \textit{et al.}~\cite{Yu12} improve upon the performance of \cite{Siddiquie11} by modelling query attribute dependency on a pool of weak attributes. Images however, are an amalgamation of several constituents, and cannot be accurately described by just a few words. It is therefore difficult to capture the complete gist of an image merely by textual descriptions \cite{Siddiquie11, Yu12}, structured queries \cite{Lan12}, concept maps \cite{Xu10} or 
sketches \cite{eitz11}. Such techniques tend to focus on one or two aspects of the image while ignoring the rest. While Lan \textit{et al.}~\cite{Lan12} consider the spatial relationship between a pair of objects, they do not account for the overall geometrical layout of all the objects and the object characteristics. Xu \textit{et al.}~\cite{Xu10} while maintaining spatial relationships do not consider background information and object attributes. Kulkarni \textit{et al.}~\cite{Kulkarni13} consider both objects and their interrelations, but do not model the background holistically.

Content based image retrieval techniques such as \cite{Douze11,Zheng14} use image queries. Zheng \textit{et al.}~\cite{Zheng14} couple complementary features of SIFT and colour into a multidimensional inverted index to improve precision, while adopting multiple assignment to improve recall. Douze \textit{et al.}~\cite{Douze11} use attributes in combination with Fisher vectors of a query image to perform retrieval. These techniques obtain a single global representation for an image, and fail to consider the objects in the image and their local characteristics. Cao \textit{et al.}~\cite{Wei14} perform image ranking by constructing triangular object structures with attribute features. However, they fail to take into account other important aspects such as the global scene context. We compare the proposed method with the works of Douze \textit{et al.} and Cao \textit{et al.} in Sec.~\ref{sec:results}.

\section{Ranking Algorithm: Aspects to Consider}
\label{sec:prob_formulation}
People tend to look for objects in images. Therefore, visually similar images would generally contain the same objects. Consider the images in Fig.~\ref{fig:eg1}. All the three images, contain the same objects, namely a person and a dog. Yet they are semantically very different from each other. This illustrates that mere presence of similar objects is insufficient to make two images similar. 

On the other hand, in both Fig.~\ref{fig:eg2} (b) and (c), not only are a dog and person present, but they are also in similar environments as in the query image. Both the retrieved images depict indoor scenarios with the objects in close spatial proximity as in the query. However, these images also contain additional objects such as a food carton, mobile and remote which are not present in the query. Also, the newspaper in Fig.~\ref{fig:eg2} (a) is missing in \ref{fig:eg2} (c). Yet, an image search returning such results would be far more appealing to a user than one returning images in Figs. \ref{fig:eg1} (b) and (c). This indicates that humans do not treat all the components of the scene with equal importance. We discuss this in more detail in Sec.~\ref{sec:node_weights}.

\begin{figure*}[t]
 \begin{minipage}{0.5\linewidth}
 \begin{center}
  \includegraphics[scale=0.43]{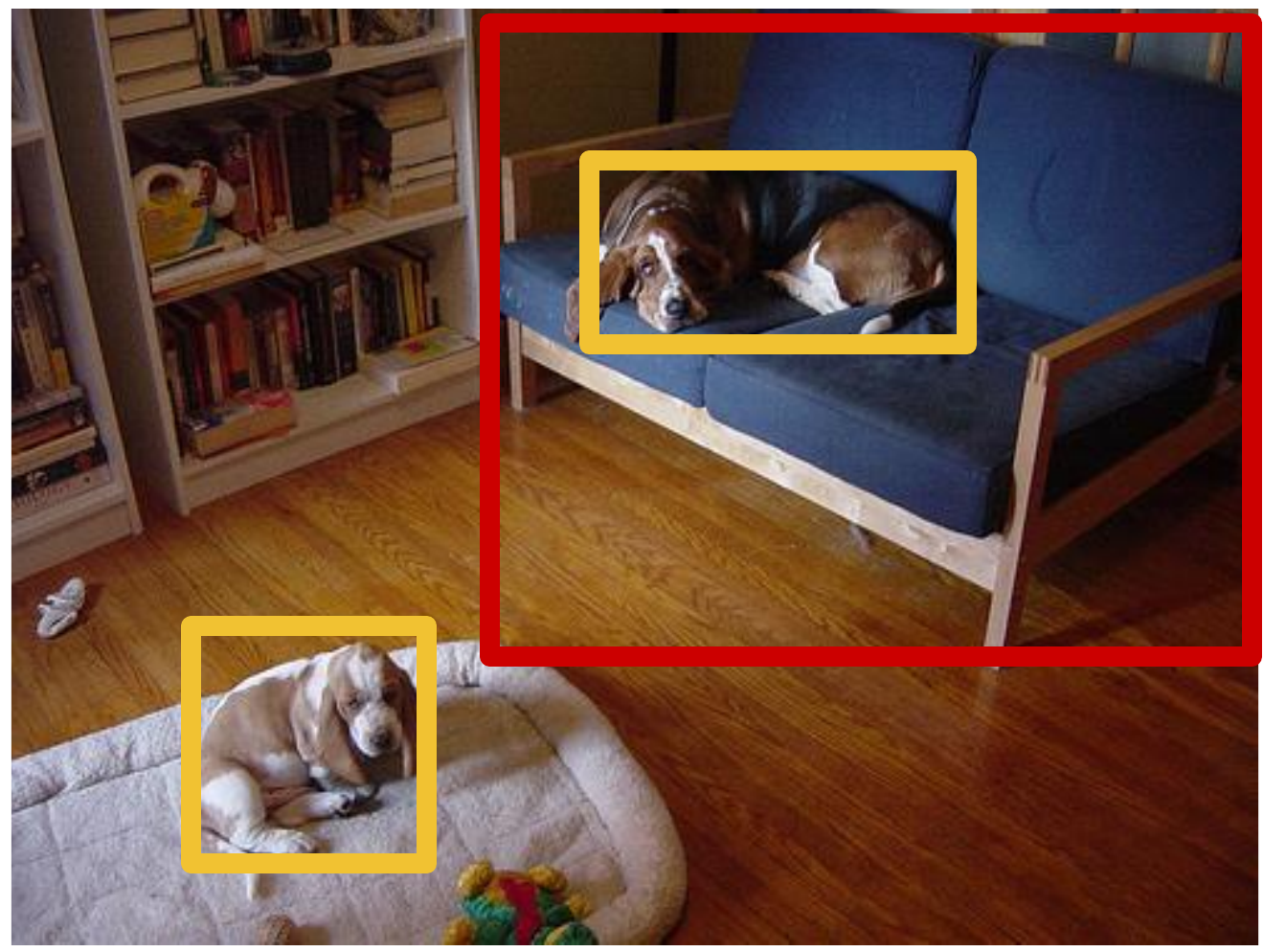}
 \end{center}
  \centerline{(a)}
 \end{minipage}
 \begin{minipage}{0.4\linewidth}
  \includegraphics[scale=0.28]{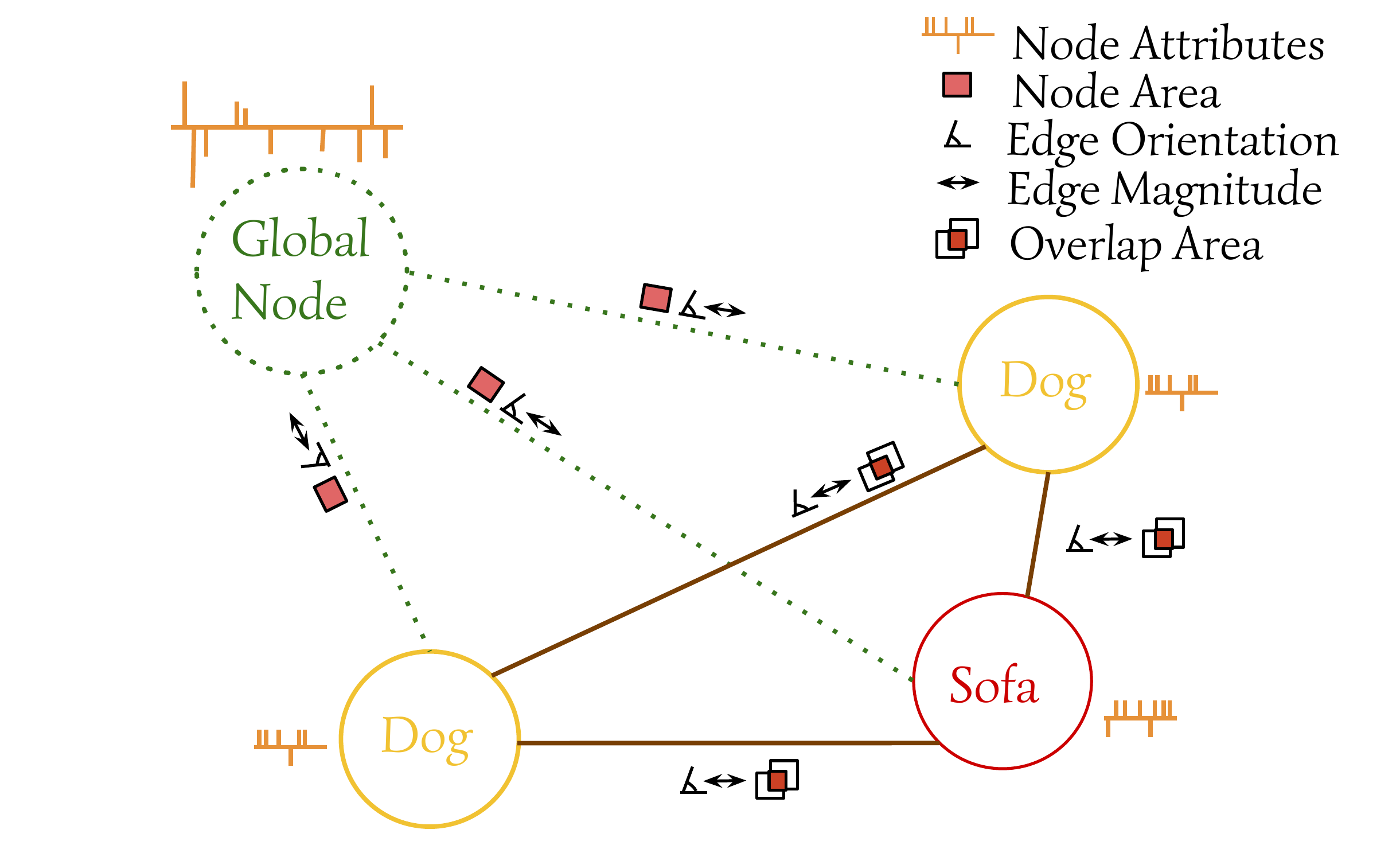}
  \centerline{(b)}
 \end{minipage}
 \caption{\footnotesize{(a) An image (b) The corresponding Attribute-Graph: The local nodes (dogs, sofa) are described by local attributes (has head, is wooden etc.) The global node captures the overall scene context and is described by global attributes (living room, interior etc.). The edges between the local nodes are characterised by the distance, angle and the overlap between the bounding boxes of objects (dogs, sofa), thereby specifying the relationship between the two. The edges connecting the local nodes to the global node define the position of that object in the scene.}}
 \label{fig:graphEg}
\end{figure*}

People usually interpret a scene using multiple cues, which are commonly referred to as image semantics. We can broadly categorise these as,
\vspace{-0.17cm}
\begin{itemize}
 \item the important objects present in the scenes
 \vspace{-0.2cm}
 \item the characteristics of these objects such as their appearance, shape, size etc.
 \vspace{-0.2cm}
 \item the spatial layout of the various scene components
 \vspace{-0.2cm}
 \item the context or background
\end{itemize}
\vspace{-0.17cm}
A ranking algorithm would need to utilise an image representation which takes all these aspects into account, in order to mimic human perception. We describe such a representation in Sec.~\ref{sec:Graph}. 

\section{Attribute-Graphs for Image Ranking}
\label{sec:proposed_method}
\subsection{Object detection and classification}
\label{sec:detect}
Objects play an important role in our understanding of an image. Therefore, accurate detection and classification of the objects is essential to construct a good representation for an image. Convolutional Neural Network (CNN) based algorithms have recently shown an improved performance over most other techniques for the tasks of detection and classification of objects~\cite{Deng09,Girshick14,Krizhevsky12}. We employ the algorithm of Girshick \textit{et al.}~\cite{Girshick14} to localise and classify objects. Girshick \textit{et al.} obtain region proposals for an image by performing selective search as described by Uijlings \textit{et al.}~\cite{Uijlings13}. A high-capacity CNN is then utilised to obtain 4096 dimensional features for the obtained region proposals.  Class specific linear SVMs are used to score these proposals. Given all scored regions in an image, a greedy non-maximal suppression is used to eliminate low scoring proposals and 
obtain the final object regions 
$R_1,R_2,\ldots$ etc., and 
their corresponding classes $o_1,o_2,\ldots$ etc. 

\subsection{Attribute-Graph construction}
\label{sec:Graph}
The proposed method uses undirected fully connected graphs $G(V,E)$ to represent images. Here, $V=\{V_l,V_g\}$ represent the nodes and $E$ represents the set of edges connecting the nodes. Each object present in the image contributes to a graph node resulting in a total of $N$ \textit{object nodes} or \textit{local nodes} denoted by $V_l=\{v_1, \ldots,v_N\}$. The additional node $V_g$ (also referred to as the \textit{global node}) represents the background or the overall scene. An image with $N$ objects is thus transformed into a graph having $N+1$ nodes. The edge sets of the graph are defined as follows: 
\begin{itemize}
 \item Local edges: Edges between two object nodes. There will be $\dbinom{N}{2}$ such edges.
 \vspace{-0.2cm}
 \item Global edges: Edges between the object nodes and the global node ($V_g$). $N$ such edges exist.
\end{itemize} 
The following sections (\ref{sec:node_features} \& \ref{sec:edge_features}) describe the features used to characterise the Attribute-Graph.
\vspace{-0.25cm}
\subsubsection{Node features}
\label{sec:node_features}
Each object node is represented using object level attributes (\textit{local attributes}). These local attributes are limited to the area occupied by the bounding box of that particular object and represent the semantic characteristics of that object. The global node captures the overall essence of the image. We use a different set of attributes, which we refer to as \textit{global attributes} to define this node. These attributes are extracted from the entire image ($I$) and describe the image as a whole. This enables the global node to represent the image context or scene characteristics effectively. The node features are assigned according to Eq.~(\ref{eq:node_feature}).

\begin{equation}
 \label{eq:node_feature}  
 \phi(v_j) = \left\{
  \begin{array}{l l}
    \varphi(I) & \quad \text{if~}v_j = V_g\\
    \psi(v_j) & \quad \text{otherwise}
  \end{array} \right. 
\end{equation}

$v_j$ represents the \textit{$j^{th}$} node, $\psi$ extracts the local attributes of the object and $\varphi$ is a function which returns the global attributes of the image. In Fig.~\ref{fig:graphEg}, the object nodes dogs and sofa are described by local attributes such as \textit{has head, is wooden} etc., and the scene by global attributes such as \textit{living room, interior} etc. 
\vspace{-0.25cm}
\subsubsection{Edge features}
\label{sec:edge_features}
The edge features of the model are defined so as to capture the spatial configuration of the image components. The local edges capture the relative arrangement of the objects with respect to each other while the global edges define the positioning of the objects in the image. The edge features are represented by Eq.~(\ref{eq:edge_feature}).
\vspace{-0.2cm}
\begin{equation}
 \label{eq:edge_feature}  
 \!\!\! \chi(e_{ij}) = \left\{\!\!\!
  \begin{array}{l l l}
    [\mu_{ij}, \theta_{ij}, o_{ij}] &  \text{if~}v_i \textit{, } v_j \in V_l\\ 
    {} & {} \\ \relax
    [\mu_{ig}, \theta_{ig}, area(v_i)] & \!\!\! \begin{array}{l}
                                         \text{if~} v_i \in V_l \:\:\&\\
                                         v_j = V_g
                                        \end{array}
  \end{array} \right.
\end{equation}

$e_{ij}$ represents the edge connecting node $v_i$ to node $v_j$. $\mu_{ij}$ is the pixel distance between object centroids. $\theta_{ij}$ represents the angle of the graph edge with respect to the horizontal taken in the anti-clockwise direction, while ensuring left/right symmetry. It indicates the relative spatial organisation of the two objects. Left/right symmetry is ensured by considering an angle $\theta$ to be equal to the angle  ($180-\theta$). $o_{ij}$ represents the amount of overlap between the bounding boxes of the two objects and is given by Eq.~(\ref{eq:overlap_bbox}).

\vspace{-0.2cm}
\begin{equation}
\label{eq:overlap_bbox}
 o_{ij} = \frac{area(v_i)\cap area(v_j)}{min(area(v_i),area(v_j))}
\end{equation}
$area(v_i)$ is the fraction of the image area occupied by the $i^{th}$ bounding box. The intersection of the two bounding boxes is normalised by the smaller of the bounding boxes to ensure an overlap score of one, when a smaller object is inside a larger one. Inclusion of $area(v_i)$ as a global edge feature causes the graph matching algorithm to match nodes of similar sizes to each other. 

$\mu_{ig}$ and $\theta_{ig}$ are the magnitude and orientation of the edge connecting the centroid of the object corresponding to node $i$ to the global centroid. The global centroid is computed as given in Eq.~(\ref{eq:gl_centroid}). 
\vspace{-0.3cm}
\begin{equation}
 \label{eq:gl_centroid}
 c_g = \frac{1}{N}\sum \limits_{k=1}^{N} c_k
\end{equation}
$c_k$ represents the centroid of the $k^{th}$ local node. The global centroid represents the centre of the geometrical layout of the objects in the image. The edges connecting each object to the global node illustrate the placement of that object with respect to the overall object topology.

The proposed algorithm to construct a graph for a given image is presented in Algorithm \ref{algo:makeGraph}. An example graph for an image is shown in Fig.~\ref{fig:graphEg}. 

\begin{algorithm}
 \SetKwInOut{Input}{Input}\SetKwInOut{Output}{Output}
 \Input{Image~(I)}
 \Output{Bounding boxes ($R_1,R_2,\ldots,R_N$), object classes ($o_1,o_2,\ldots,o_N$), nodeFeature, edgeFeature}
  \BlankLine
  /* N - no. of objects */ \\
 {[$R_1,R_2,\ldots,R_N$]} = objectDetectors(I)\;
 [$o_1,o_2,\ldots,o_N$] = objectClassifiers($R_1,R_2,\ldots,R_N$);
 \BlankLine
 /* Extract node features */\\
 \For {i = 1:N} 
  {
   nodeFeature($v_i$) = extractLocalAttribute(I, $R_i$)\;
  }
  nodeFeature($v_{g}$) = extractGlobalAttribute(I);\\
  \BlankLine
  \For {j = 1:N}
   {
     \For {k = j+1:N}
      {
        /* Construct local edges and extract their features */\\
        $e_{jk}$ = getEdge($v_j$, $v_k$)\;
        edgeFeature($e_{jk}$) = [$\mu_{jk}$, $\theta_{jk}$, $o_{jk}$]\;
      }
   }
   $c_g$ = centroid($c_1,c_2,\ldots,c_N$);\\ /*$c_1,c_2,\ldots\rightarrow$ centroids of $R_1,R_2,\ldots$*/\\
   \BlankLine
  \For {p = 1:N}
   {
     $e_{pg}$ = getEdge($v_p$, $c_g$); /* Construct global edges */\\
     edgeFeature($e_{pg}$) = [$\mu_{pg}$, $\theta_{pg}$, $area(v_p)$]\;
   }
  
 \caption{Construct Image Attribute-Graph}
 \label{algo:makeGraph}
\end{algorithm}

\begin{algorithm}
 \SetKwInOut{Input}{Input}\SetKwInOut{Output}{Output}
 \Input{Query Image (Q)}
 \Output{Ranklist}
 \BlankLine
 /* Construct query Attribute-graph using Algo.~(\ref{algo:makeGraph}) */ \\
 {[qGraph, bounding boxes ($R_1,R_2,\ldots,R_N$), object classes ($o_1,o_2,\ldots,o_N$)]} = constructGraph(Q)\;  
 $wts$ = getWeight($R_1,R_2,\ldots,R_N$)\; /* Weights indicate the relative importance of different query objects */ \\
 \BlankLine
 \For{image \textnormal{in} Dataset images}
 {
 \BlankLine
  /* load Dataset image graph */ \\
  dGraph = loadGraph(image)\;
  [$S_{lcl}$, $S_{gbl}$, $S_{edge}$] = graphMatch(qGraph, dGraph, $wts$, ($o_1,o_2,\ldots,o_N$))\;
  /* Calculate score */ \\
  $score = \alpha \times S_{lcl} + \beta \times S_{gbl} + (1-\alpha-\beta) \times S_{edge}$
 }
 \BlankLine
 Ranklist = sort(score);
 \caption{Attribute-Graph Ranking}
 \label{algo:ranking}
\end{algorithm}

\subsection{Relative Importance of different objects}
\label{sec:node_weights}
As discussed in Sec.~\ref{sec:prob_formulation}, humans do not attach equal importance to all objects. The presence/absence of relatively smaller objects such as food carton, mobile and remote do not seem to make much of a difference in our perception of the images in Fig.~\ref{fig:eg2}. However, those images would have seemed significantly different to us, had the relatively larger objects (person/dog) been missing. This indicates varying importance for objects in the scene depending on the their attributes. Further, Proulx \cite{Proulx10,Proulx11} perform visual search experiments to conclude that larger objects capture greater visual attention.

We deal with this unequal importance of different objects, by assigning a relative \textit{weight} to each local node of the query. This weight is computed as the ratio of the area of the corresponding bounding box to the sum total of all bounding box areas. This causes larger objects to be given priority during matching. 

\subsection{Graph Matching}
\label{sec:GM}
During the evaluation phase, we use graph matching to compare the query image  Attribute-Graph with each of the dataset image Attribute-Graphs. Graph matching is done such that the overall similarity score between the mapped nodes and edges is maximised, while preserving the matching constraints. We enforce the following constraints on the matching process:
\vspace{-0.2cm}
\begin{itemize}
 \item A local node of a particular object class can be matched only to another local node of the same class.
 \vspace{-0.3cm}
 \item A global node can be matched only to another global node.
\end{itemize}

For our experiments, we use Re-weighted Random Walks (RRWM) algorithm proposed by Cho \textit{et al.}~\cite{Cho10}, which formulates the problem of matching as the task of node selection on an association graph. This graph is constructed by modelling the nodes as the candidate correspondences between the graphs to be matched. Random walks are then performed on the association graph, with intermediate re-weighting jumps, while enforcing the matching constraints. This graph matching algorithm has been shown to be robust to noise, outliers and deformation. However, our Attribute-Graph representation is general and any other graph matching technique, capable of handling constraints, can also be used to obtain matching scores between two Attribute-Graphs.

The scores of each dataset image are given by Eq.~(\ref{eq:graph_score}).
\vspace{-0.3cm}
\begin{equation}
\label{eq:graph_score}
  [S_{lcl}, S_{gbl}, S_{edge}] = GM(Q,D_k,wts,objs) 
  \vspace{-0.1cm}
\end{equation}
where $Q$ refers to the query image, $D_k$ is the $k^{th}$ image in the dataset \textit{D} and \textit{GM(.,.)} is the graph matching technique. $S_{lcl}$, $S_{gbl}$, $S_{edge}$ are the scores associated with the matching between the local nodes, the global node and the edges respectively of the two graphs. $wts$ correspond to the importance weights calculated as described in Sec.~\ref{sec:node_weights}. $objs$ refers to the object classes associated with the local nodes.

The algorithm for obtaining a ranked set of images given a query image is described in Algorithm \ref{algo:ranking}. The final score corresponding to a dataset image is calculated as shown in algorithm. $\alpha$ and $\beta$ are constants determined empirically. The scores obtained are sorted to get the ranklist. An overview of the proposed method is depicted in Fig. \ref{fig:Overview}.

\section{Datasets}
\label{sec:dataset}
To evaluate our ranking technique, we require datasets which have graded relevance scores for the reference images corresponding to each query. Moreover, it is essential that these relevance scores have been assigned considering overall visual similarity, and not just one particular component of the image (For e.g. according to ImageNet hierarchy which focuses mostly on objects). For this purpose we have created two datasets, \textit{rPascal} and \textit{rImageNet} which are subsets of the aPascal~\cite{Farhadi09} and ImageNet~\cite{imagenet09} databases. Each of these datasets contain 50 query images and a set of reference images corresponding to each query. Reference images correspond to the set of images that are considered for ranking for a given query image. These reference images have been selected so as to contain at least one object in common with that particular query. This has been done to prevent burdening the annotators with a lot of irrelevant images.

\begin{figure*}
 \begin{minipage}{0.49\linewidth}
 \begin{center}
\includegraphics[scale=0.16]{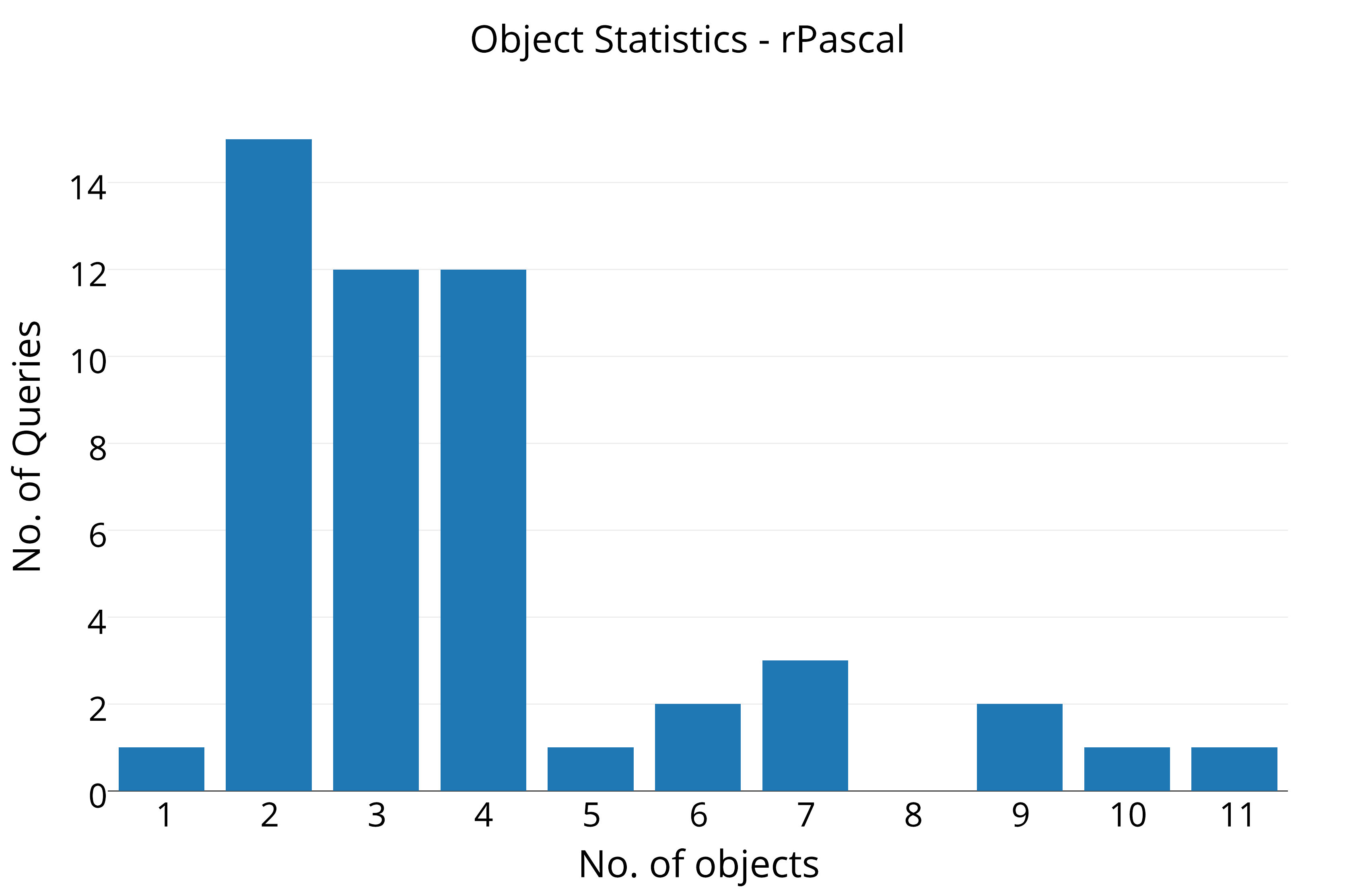}
 \end{center}
 \centerline{(a)}
\end{minipage}
\begin{minipage}{0.49\linewidth}
\begin{center}
\includegraphics[scale=0.16]{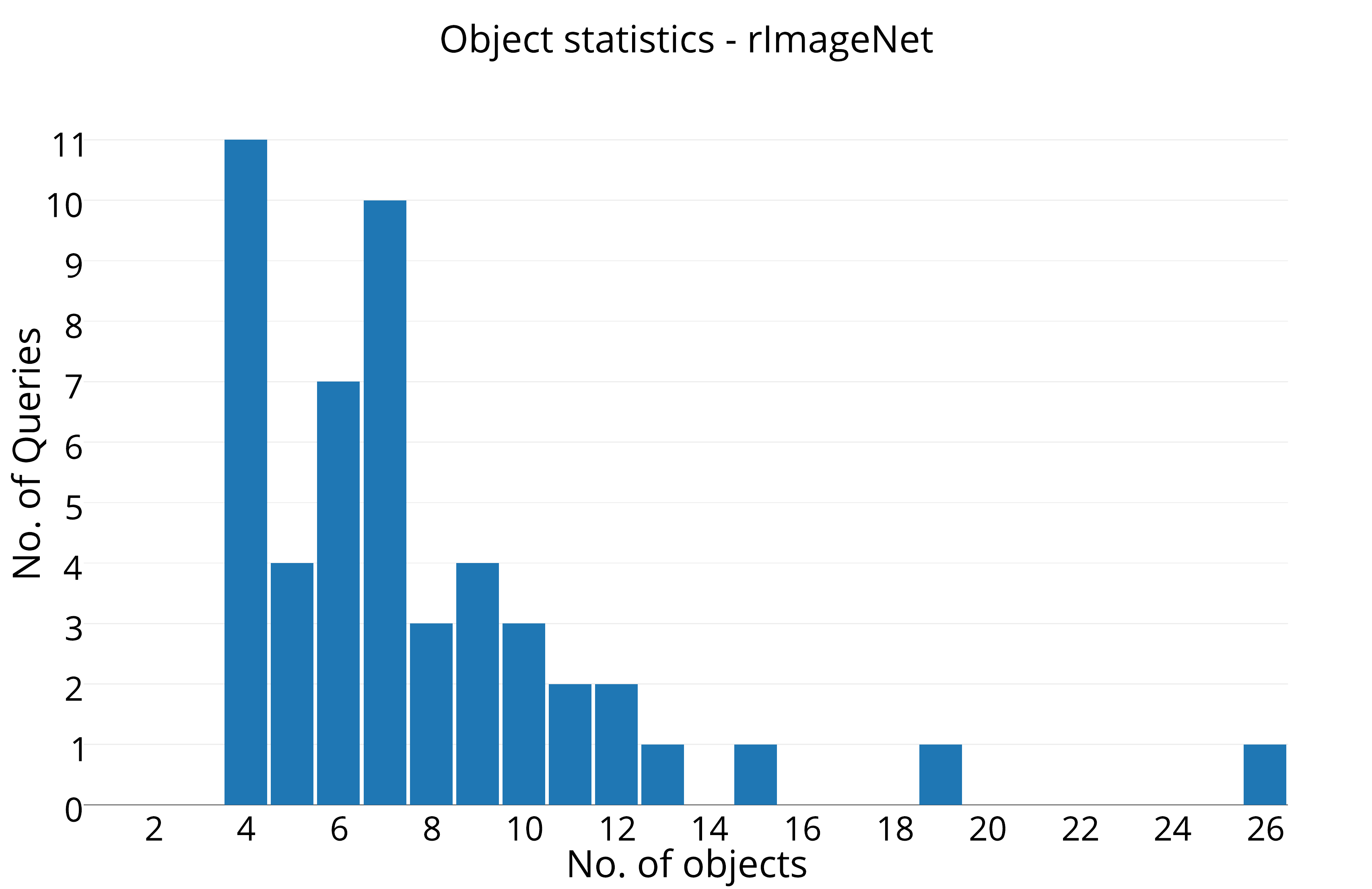}
\end{center}
 \centerline{(b)}
\end{minipage}
 \caption{Object count in query images of (a)~\textit{rPascal} (b)~\textit{rImageNet}}
 \label{fig:Dataset}
\end{figure*}

12 volunteers were recruited to annotate the images. Given a query image and the corresponding reference images, the annotators were asked to give a relevance score between 0-3 for each of the reference images with respect to that particular query. These score values were defined as follows:
\vspace{-0.2cm}
\begin{itemize}
 \item 0 - irrelevant : Unrelated to the query and should not be retrieved when searching for this particular query
 \vspace{-0.25cm}
 \item 1 - fair : Has a few components/aspects similar to the query
 \vspace{-0.25cm}
 \item 2 - good : Similar to the query except for a few missing components/aspects
 \vspace{-0.25cm}
 \item 3 - excellent : Very similar, exactly what we would like to see when searching for this particular query
\end{itemize}
\vspace{-0.1cm}

Each annotator was presented with reference images corresponding to 20 query images from each dataset. This resulted in the reference set corresponding to each query image receiving annotations from at least 5 different annotators. The final relevance score was obtained by computing the median score of the 5 annotations for each image.
\vspace{-0.2cm}
\subsection{rPascal Dataset}
The \textit{aPascal} dataset \cite{Farhadi09} contains a total of 4340 images, spanning 20 labelled object classes, divided into 2113 training images and 2227 test images. Our \textit{`rPascal'} (ranking Pascal) dataset is constructed entirely over the test set of aPascal. The queries consist of 18 indoor scenes and 32 outdoor scenes. The queries include both simple images containing one or two objects and also more complex images containing 8 or more objects. The \textit{rPascal} dataset contains a total of 1,835 images with an average of 180 reference images per query.
\vspace{-0.2cm}
\subsection{rImageNet}
The \textit{rImageNet} dataset is constructed from the validation set of ILSVRC 2013~\cite{ILSVRC13} detection challenge, which contains 20,121 images with objects belonging to 200 different classes. To increase the complexity of the dataset, we have selected only images containing at least 4 objects. The queries contain 14 indoor and 36 outdoor scenes. The \textit{rImageNet} dataset contains a total of 3,354 images with an average of 305 reference images per query. Owing to the increased number of classes and objects in the images, this dataset is much more challenging than \textit{rPascal}. 

The statistics pertaining to the number of queries containing a certain number of objects for both the datasets is given in Fig.~\ref{fig:Dataset}. Example queries from both datasets can be seen in the first column of Fig.~\ref{fig:results_attrgraph}.

\section{Experiments and Results}
\label{sec:results}
\subsection{Experimental setup}

\begin{figure*}[!t]
 \begin{minipage}{0.5\linewidth}
  \includegraphics[scale=0.23]{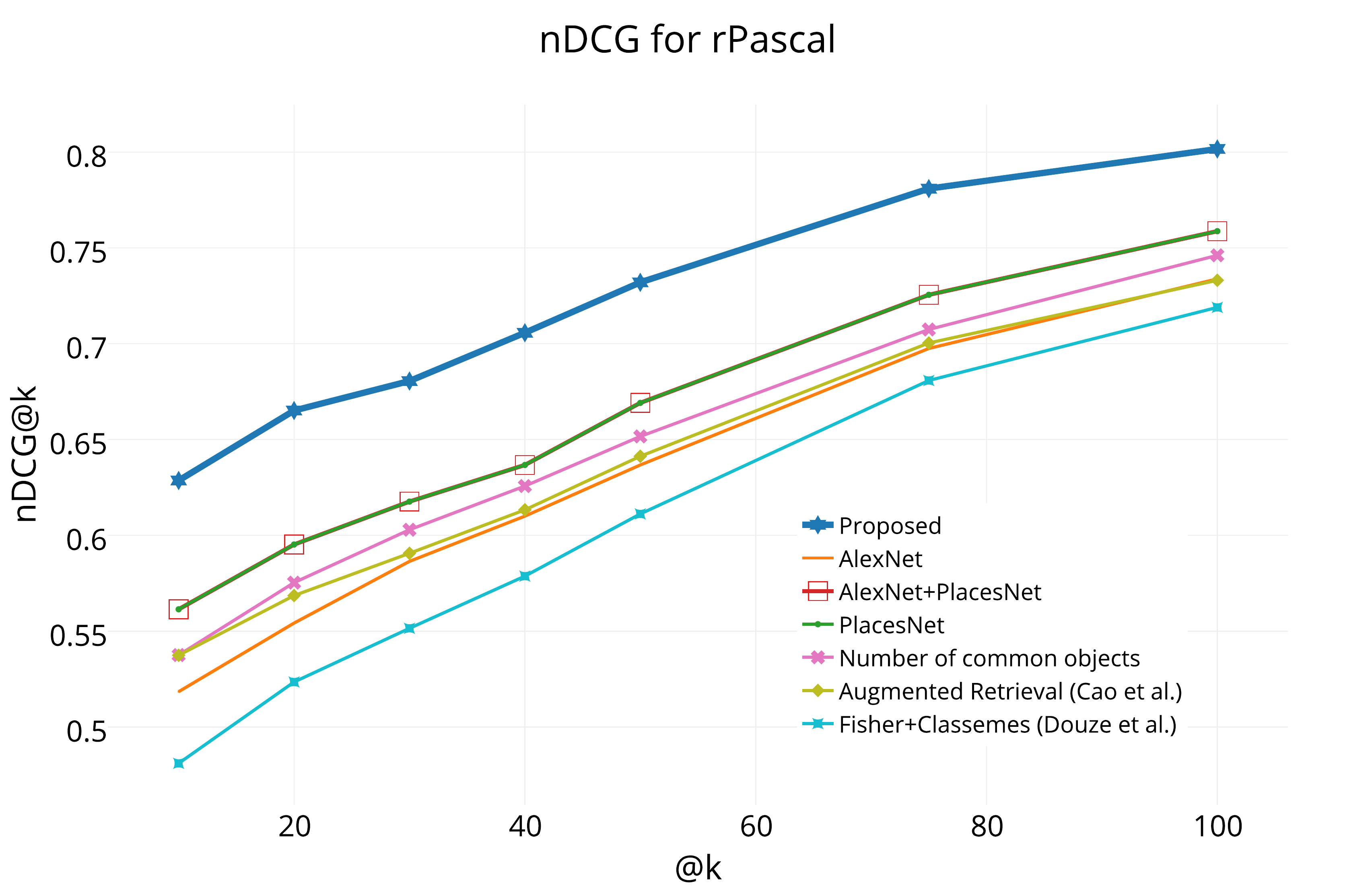}
  \centerline{(a)}
 \end{minipage}
 \begin{minipage}{0.5\linewidth}
  \includegraphics[scale=0.23]{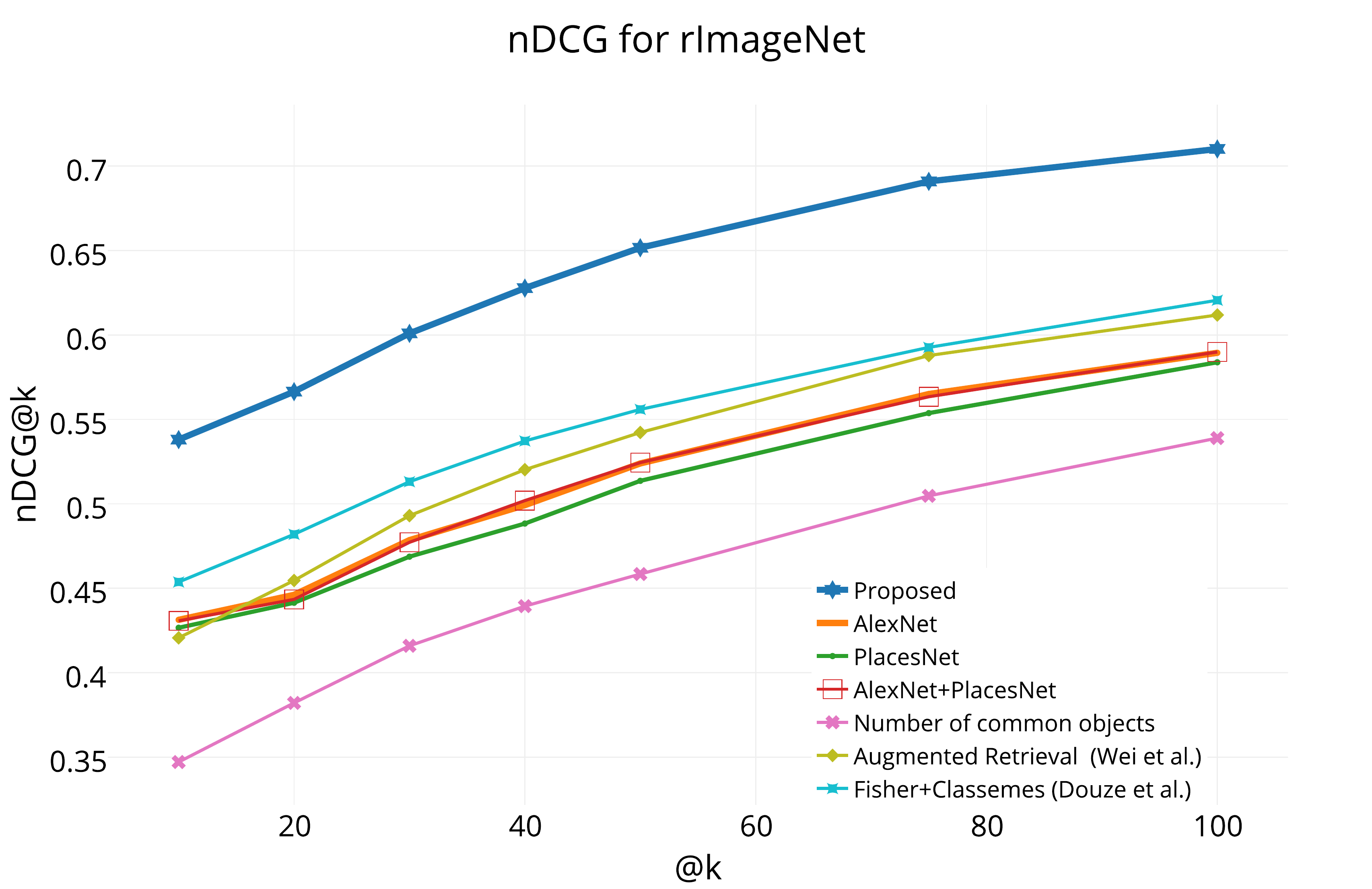}
  \centerline{(b)}
 \end{minipage}
 \caption{Qualitative results: nDCG comparison against Ranking truncation level `k' for (a) \textit{rPascal} dataset (b) \textit{rImageNet} dataset. The proposed method shows a significant improvement in performance over both Douze \textit{et al.}~\cite{Douze11} and Augmented Retrieval~\cite{Wei14} as well as the other baselines in both datasets. Plots are best viewed by zooming in.}
 \label{fig:ndcg_plots}
\end{figure*}

We perform object detection and classification, using Regions with CNN (RCNN), the algorithm of Girshick \textit{et al.}~\cite{Girshick14}. These detected regions form the nodes of our graph and are characterised by the local attributes extracted from them. We employ the 64 attributes defined by Farhadi \textit{et al.}~\cite{Farhadi09} as our local attributes. They consist of shape attributes such as `2D boxy', `cylindrical' etc., part attributes such as `has head', `has leg', `has wheel' etc. and material attributes such as `has wood' and `is furry'. We use the same set of features as those used by Farhadi \textit{et al.}~\cite{Farhadi09}, which describe each object by a 9751 dimensional feature vector containing texton, HOG, edge and colour descriptor based visual words, to train the attribute classifiers. The attribute classifiers are trained on the aPascal training set. For global attributes, we use the 205 dimensional output probabilities of a deep network trained on the Places database\footnote{The pre-trained network was obtained at http://places.csail.mit.edu/}\cite{zhou2014places}, which have classes such as coast, desert, forest, home, hotel etc. We set $\alpha=0.4$, $\beta=0.4$ (in Algo.~\ref{algo:ranking}) in all our experiments. 

For a baseline, we compare with our implementation of the work of Douze \textit{et al.}~\cite{Douze11}\footnote{Our implementation achieves the accuracies mentioned in \cite{Douze11} on the databases used by Douze \textit{et al.}}. Douze \textit{et al.} uses Fisher vectors \cite{Perronnin07} (of SIFT features) extracted from the image concatenated with `Classemes' \cite{Torresani10} as features to describe an image. Classemes is a 2659 dimensional vector of classifier outputs, trained for categories that are selected from an ontology of visual concepts. The features used to obtain the classemes are extracted from the image as a whole and consist of colour GIST, oriented and unoriented PHOG, pyramid self similarity~\cite{Shechtman07} and bag of words using the SIFT descriptor.

We also compare the proposed method with the attribute based ranking method of Cao \textit{et al.}~\cite{Wei14}. They represent an image using a set of disjoint triangles, whose vertices correspond to image objects. With this representation, they calculate the similarity between two images by mapping the triangles of one image on to the other. 

\subsection{Evaluation measures}
For quantitative evaluation, we compute the Normalised Discounted Cumulative Gain (nDCG) of our ranklist. nDCG is a standard measure for evaluating ranking algorithms~\cite{Wei14,Siddiquie11} and is given in Eq.~(\ref{eq:ndcg}). 
\vspace{-0.3cm}
\begin{align}
 \label{eq:ndcg}
  DCG@k = \sum \limits_{i=1}^{k} \frac{2^{rel_i}-1}{log_2(i)} \nonumber \\
  nDCG@k = \frac{DCG}{IDCG}
\end{align}
$rel_i$ is the relevance of the $i^{th}$ ranked image and IDCG refers to the ideal Discounted Cumulative Gain(DCG), which acts as a normalisation constant to ensure that the correct ranking results in an nDCG score of 1. $k$ denotes the \textit{ranking truncation level}.

\begin{figure*}[!t]
 \begin{minipage}{0.5\linewidth}
  \centering
  \includegraphics[scale=0.22]{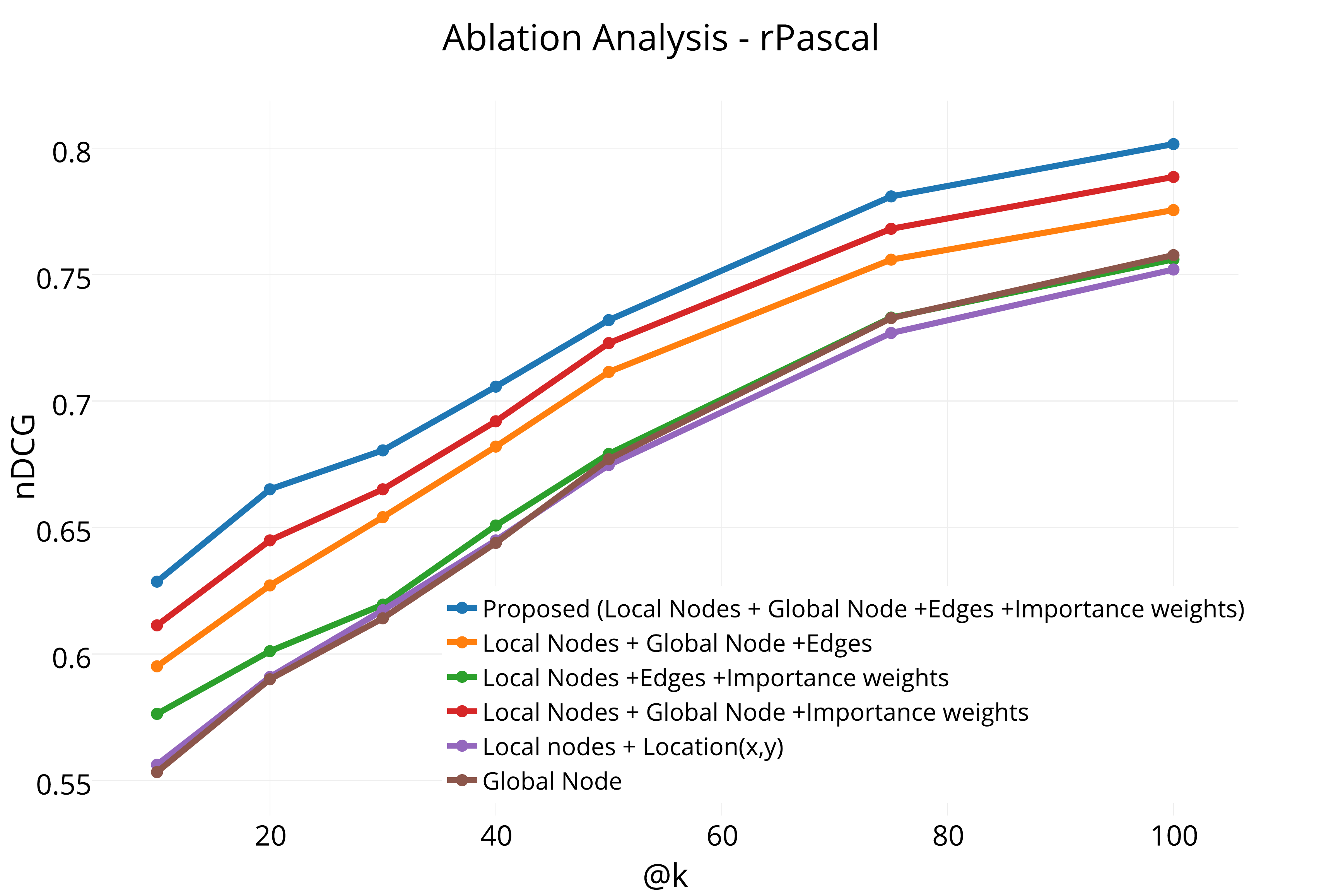}
  \centerline{(a)}
 \end{minipage}
 \begin{minipage}{0.5\linewidth}
  \centering
  \includegraphics[scale=0.22]{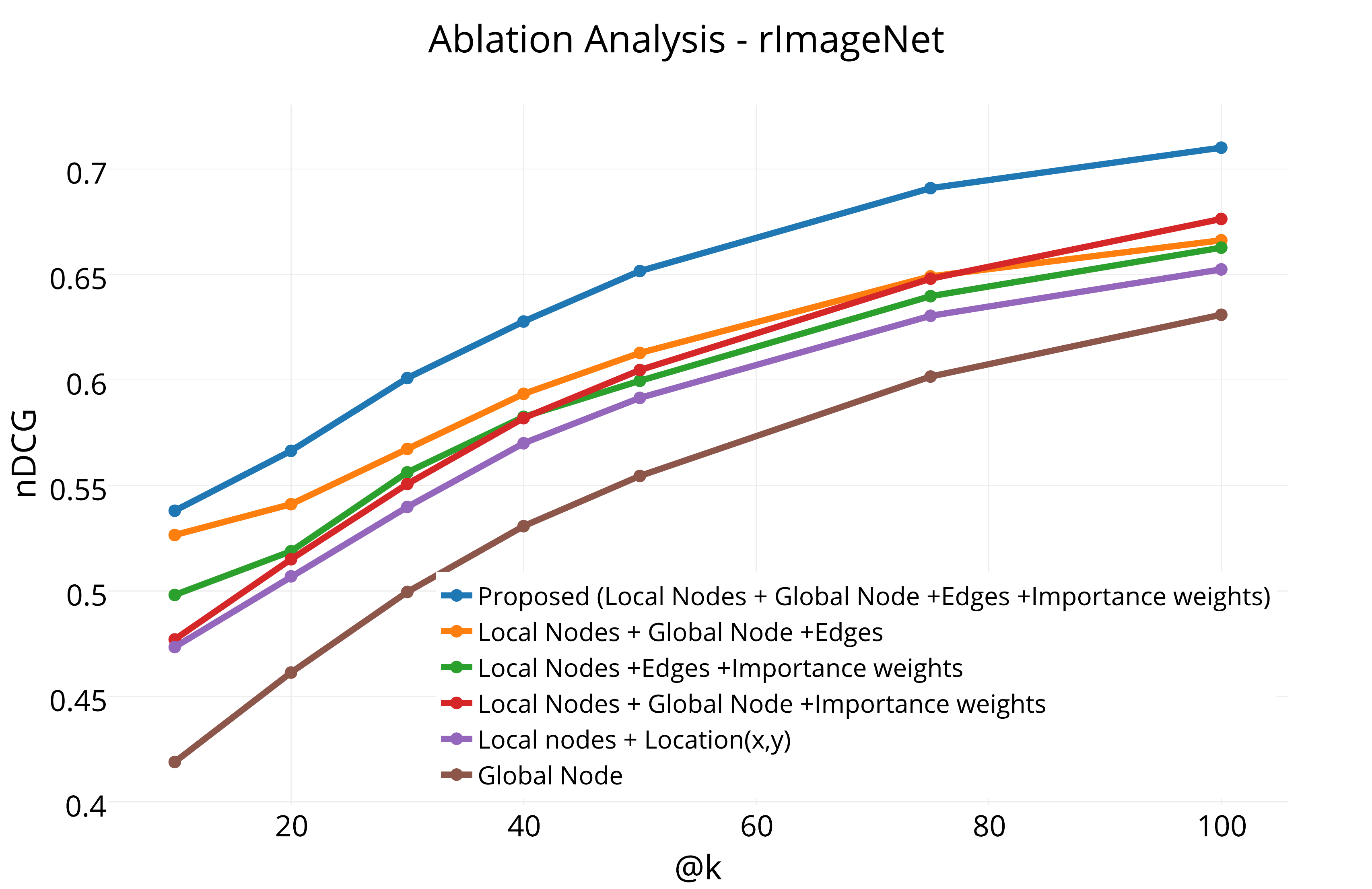}
  \centerline{(b)}
 \end{minipage}
 \caption{Ablation analysis of the proposed method for (a)~\textit{rPascal} (b)~\textit{rImageNet}. Plots are best viewed by zooming in.}
 \label{fig:ablation_plots}
\end{figure*}
\subsection{Discussion}
\label{sec:discussion}
Figure~\ref{fig:ndcg_plots} plots the nDCG scores of the proposed method, Douze \textit{et al.}~\cite{Douze11}, Cao \textit{et al.}~\cite{Wei14}, as well as a some other baselines as a function of the ranking truncation level $k$. The baselines we consider include nearest neighbour comparison with the fc7 features of AlexNet~\cite{Krizhevsky12}, PlacesNet~\cite{zhou2014places}, and a combination of the two. We also compare against a simple baseline which ranks images based on just the number of common objects present in the two images. As observed from the figure, our performance exceeds that of the other techniques~\cite{Wei14,Douze11} as well as the above established baselines on both datasets. Since Douze \textit{et al.} consider only the global descriptors of images and do not try to model the objects and their spatial layout, their results in many cases, fail to have 
all the desired objects in the first few retrieved images.

Cao \textit{et al.} not only ignore the global scene context, but also do not characterise the relationships existing between their object triangles, thereby limiting the amount of spatial layout information up to three objects. As can be observed from Fig.~\ref{fig:ndcg_plots}, these drawbacks lead to a poor performance, especially when the queries are complex images with many objects as in the case of the \textit{rImageNet} dataset.

\begin{figure*}[!t]
 \centering
\includegraphics[scale=0.59]{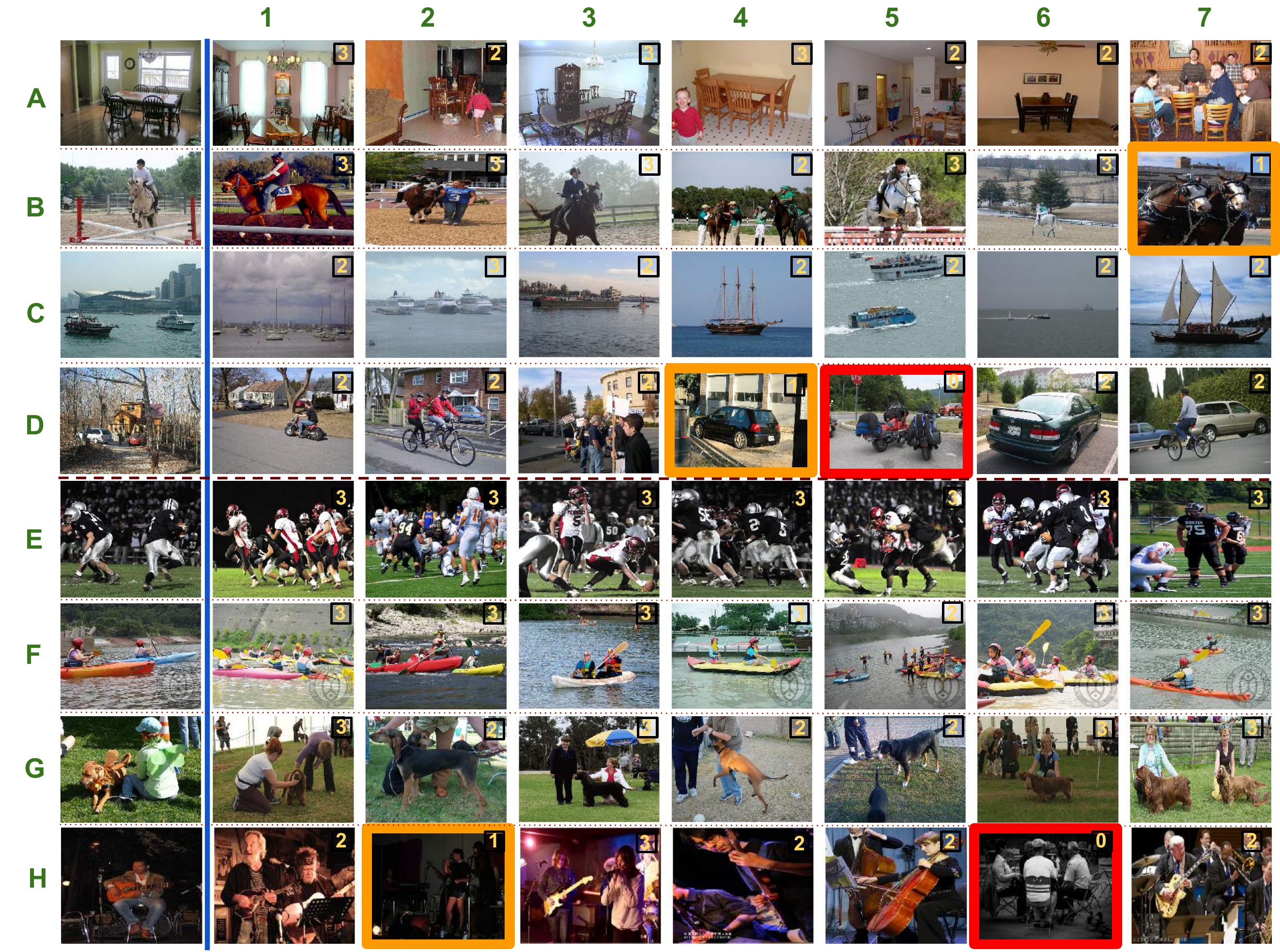}
 \caption{Ranking results for the Proposed method: Column 1: Queries A-H, Columns 2-8: First 7 retrieved images. The first 4 queries (A-D) are from \textit{rPascal} and the last 4 (E-H) belong to \textit{rImageNet}. The annotation scores are shown at the right top corner of every image. \textit{Irrelevant} images have been marked with a red boundary and images with an annotation of \textit{Fair} have been marked with orange. The remaining images have been annotated as \textit{Good} or \textit{Excellent}.}
 \label{fig:results_attrgraph}
\end{figure*}

On \textit{rPascal} we show an improvement of 5-8\% over Cao \textit{et al.} and 7-12\% over Douze \textit{et al.} On \textit{rImageNet} we show an improvement of 9-11\% over Cao \textit{et al.} and 8\% over Douze \textit{et al.}  

Figure~\ref{fig:results_attrgraph} shows a few ranking examples by the proposed method. The images ranked 1-7 for each of the queries in the first column are shown in columns 2-8 of Fig.~\ref{fig:results_attrgraph}. Ranked images annotated as \textit{fair} and \textit{irrelevant} have been marked with orange and red boxes respectively. Two of these appear in queries corresponding to rows \textit{D} and \textit{H}. This mis-ranking is mainly due to incorrect detections and classifications of objects in the reference images. For example, the two bikes in row \textit{D}, column 6, are wrongly classified as cars. Similarly, the $6^{th}$ ranked image in the last row has mis-detections of microphone.\\ 
The proposed method scales linearly with the number of images in the database. An inverted index file scheme based on objects and attributes can be used to speed this up further.
\vspace{-0.5cm}
\subsection{Ablation Analysis}
\vspace{-0.15cm}
To better understand the contribution of various components of the Attribute-Graph, we analysed the performance of the proposed method, by ablating each component. The results obtained are depicted in Fig.~\ref{fig:ablation_plots}. Our analysis confirms that removal of any of the global node, object nodes, edges or importance weights assigned to the nodes, negatively impacts performance, thereby demonstrating the importance of each of the components of the proposed Attribute-Graph. We observe that the  performance is most affected by the removal of object nodes leading to 7\% and 12\% drops in \textit{rPascal} and \textit{rImageNet} respectively.

\vspace{0.1cm} 
\section{Conclusion}
\label{sec:conclusion}
\vspace{-0.1cm}
We have proposed a novel image representation using an attribute based graph structure. Our Attribute-Graphs represent various objects present in the image, along with their characteristics such as shape, texture, material and appearance. Attribute-Graphs also capture the spatial scene structure via the graph edges and the overall scene gist through global attributes. We show the efficacy of the proposed representation by its application to image ranking. We evaluate the performance of our ranking technique on the \textit{rPascal} and \textit{rImageNet} datasets, which we have collated for the purpose. The proposed method obtains an improvement of around 5-8\% in the nDCG scores, demonstrating the ability of our representation in capturing semantic similarity.

\section{Acknowledgements}
\vspace{-0.25cm}
This work was supported by Defence Research and Development Organization (DRDO), Government of India.
Supported by Microsoft Research India Travel Grant.

{\footnotesize
\bibliographystyle{ieee}
\bibliography{egbib}
}

\end{document}